\documentclass[10pt,a4paper]{article}
 
\usepackage[T2A]{fontenc}
\usepackage[utf8]{inputenc}
\usepackage{amsmath}
\usepackage{amsfonts}
\usepackage{amssymb}
\usepackage[style=numeric,backend=bibtex,sorting=none]{biblatex}
\usepackage{algorithm}
\usepackage{algpseudocode}
\usepackage{graphicx}
\usepackage{subfigure}
\usepackage[a4paper]{geometry}
\usepackage{adjustbox}
\usepackage{graphicx}
\usepackage{rotating}
\usepackage{multirow}

\begin{document}

\title{Neuromorphic Artificial Intelligence Systems}
\author{
Dmitry Ivanov\\
\textit{rudimiv@gmail.com}\\
Cifrum, Lomonosov Moscow State University \and 
Aleksandr Chezhegov\\
\textit{chezhegov@nanolab.phys.msu.ru}\\
Lomonosov Moscow State University \and
Andrey Grunin\\
\textit{grunin@nanolab.phys.msu.ru}\\
Lomonosov Moscow State University \and
Mikhail Kiselev\\ 
\textit{mkiselev@chuvsu.ru}\\
Cifrum, Chuvash State University \and
Denis Larionov\\
\textit{denis.larionov@gmail.com}\\
Cifrum
}
\date{May 2022}
\maketitle

\begin{abstract}
Modern AI systems, based on von Neumann architecture and classical neural networks, have a number of fundamental limitations in comparison with the brain. This article discusses such limitations and the ways they can be mitigated. Next, it presents an overview of currently available neuromorphic AI projects in which these limitations are overcame by bringing some brain features into the functioning and organization of computing systems (TrueNorth, Loihi, Tianjic, SpiNNaker, BrainScaleS, NeuronFlow, DYNAP, Akida). Also, the article presents the principle of classifying neuromorphic AI systems by the brain features they use (neural networks, parallelism and asynchrony, impulse nature of information transfer, local learning, sparsity, analog and in-memory computing). In addition to new architectural approaches used in neuromorphic devices based on existing silicon microelectronics technologies, the article also discusses the prospects of using new memristor element base. Examples of recent advances in the use of memristors in neuromorphic applications are also given.

\end{abstract}

\newpage
\section{Introduction}

Modern artificial intelligence (AI) systems based on neural networks would not be possible without hardware that can quickly perform a huge number of repetitive parallel operations. The modern AI systems have become possible and gained widespread use due to hardware and large datatasets. As shown in \cite{hooker2020hardware}, throughout the AI history, precisely those approaches won for which there was suitable hardware. That is why it is important to consider \textit{AI algorithms in conjunction with the hardware that they are run on}. It is the hardware that determines the availability and effectiveness of AI algorithms.

A great majority of the latest AI systems are built by pairing von Neumann computers and classical neural networks, dating back to the Rosenblatt’s perceptron.

\subsection{Von Neumann Architecture}

The von Neumann architecture separates the memory and the computations. The computations are executed in the form of programs, which are sequences of machine instructions. Instructions are performed by a processor. A processor instruction usually has several arguments that it takes from processor registers (small but very fast memory cells located in the processor). At that, the instructions and most of the data are stored in the memory separately from the processor. The processor and the memory are connected by a data bus by which the processor receives instructions and data from the memory.

The first bottleneck of this architecture is the limited throughput of the data bus between the memory and the processor. The data bus is loaded mainly by transfer of intermediate calculation data from/to random access memory (RAM) during the execution of a program. Moreover, the maximum throughput of the data bus is much less than the speed at which the processor can process data.

Another important limitation is the largely different speed of the RAM and the processor registers (Figure 1). This can cause latency and processor downtime while it fetches data from the memory. This phenomenon is known as the von Neumann bottleneck.

It is also worth noting that this approach is energy-intensive. As argued in \cite{horowitz2014}, the energy needed for one operation of moving data along the bus can be 1,000 times more than the energy for one computing operation. For example, adding two int8s consumes \~0.03 pJ while reading from DRAM consumes \~2.6 nJ.

\subsection{Neural Networks based on the Von Neumann Architecture}

To solve cognitive problems on computers, there were developed the concept of artificial neural networks (ANN) based on the perceptron \cite{dlbook} and the backpropagation method \cite{rumelhart1986learning}. Perceptron is a simplified mathematical model of an artificial neuron network, in which neurons calculate a weighted sum of their input signals and generate an output signal using an activation function. The process of training such a network by the backpropagation method consists in modification of its weights decreasing the error. 

Since the majority of modern neural networks have a layered architecture, the most computationally intensive operation in these networks is the operation of multiplying a matrix by a vector $y = Wx$. To carry out this operation, it is first necessary to obtain data from the memory, namely, $ m * n $ weights of $W$ and $n$ values of vector $x$. It should be noted that, $m * n$ weights will be used once per matrix-vector multiplication while the values from the vector $x$ will be reused (Figure 2).

Thus, in order to perform computations, the processor needs to receive weights and input data from the memory. As mentioned above, the throughput of the data bus and the latency in receiving data limit the speed of obtaining weights. Also, the number of weights grows as $O(n^2 )$, where n is the input size. However, there are the limit of the throughput of the data bus, connecting the processor with the memory and transferring the weights and the input vectors. It becomes exhausted much earlier than the available amount of computation (multiplying matrices by a vector and taking activation functions) per unit time.

\begin{figure}
  \includegraphics[width=\linewidth]{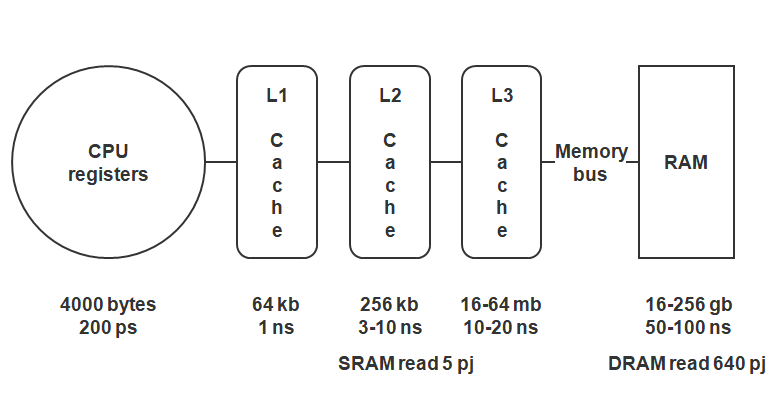}
  \caption{Memory hierarchy, access speed and power consumption}
  \label{fig:memory-hierarchy}
\end{figure}

\begin{figure}
  \includegraphics[width=\linewidth]{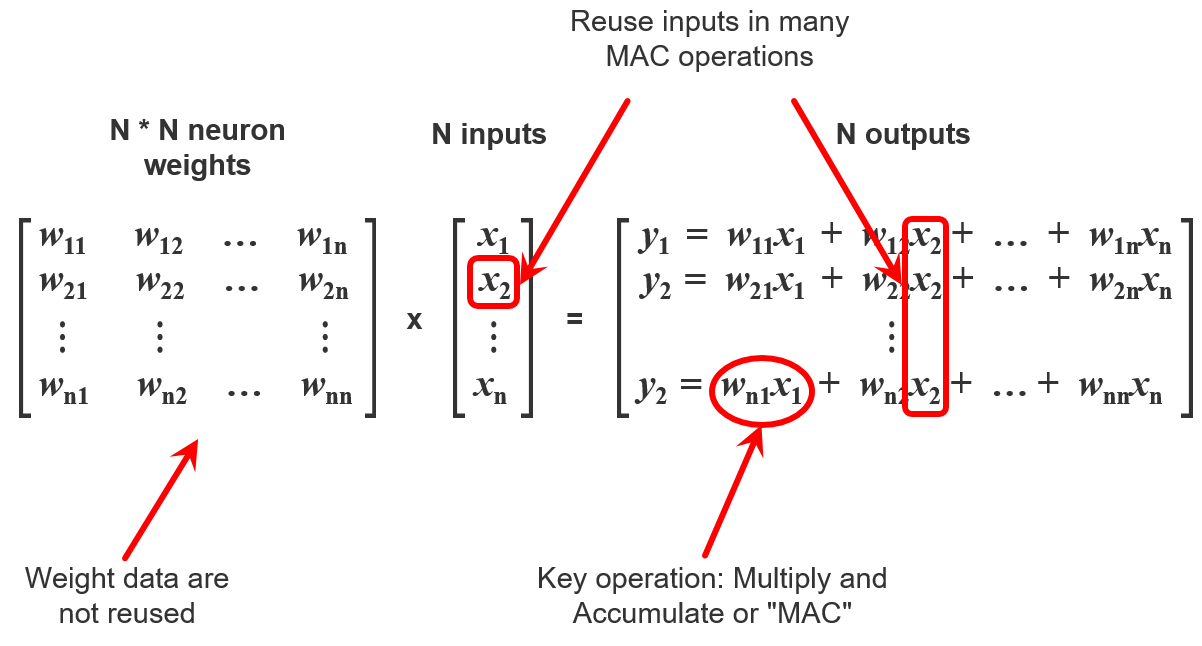}
  \caption{A schematic view of neural network computations. Elements of input vector x are reused n times while weights \(w_{ij}\) are used once}.
  \label{fig:matrix-mul}
\end{figure}

\subsection{Mitigating Limitations in Modern Computing Systems}

To begin with, let us look at the ways to mitigate the above limitations in modern AI systems.

\subsubsection{CPU}
\begin{figure}
  \includegraphics[width=\linewidth]{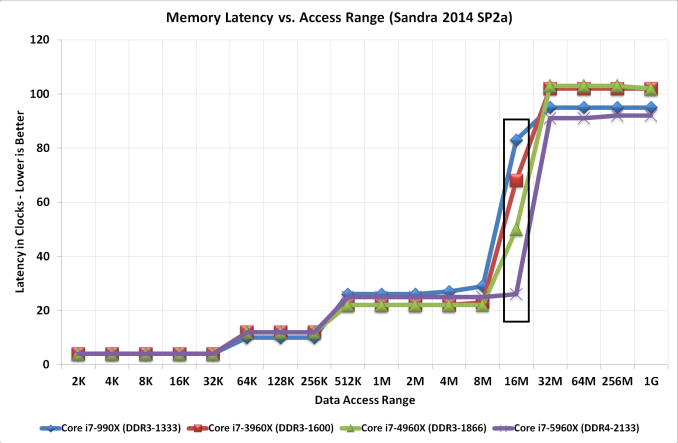}
  \caption{Memory latency depending on data range \cite{loihi_arm}}
  \label{fig:memory-latency-vs-size}
\end{figure}

Classically, the problem of memory latency was solved in the CPU by using a complex multi-level cache system \ref{fig:memory-hierarchy} \cite{hennessy2019computer}. In modern processors, caches can be up to 40\% of the chip area, providing tens of megabytes of ultra-fast memory. Usually, the size of practically used neural networks does not allow fitting all the weights into caches (see \ref{fig:memory-latency-vs-size}). Nevertheless, the latest processors with AMD 3D V-cache technology can change this situation, providing larger caches.

Other traditional approaches to optimizing processors were speculative execution, branch prediction, and others \cite{hennessy2019computer}. However, in matrix multiplication, the order of computations is known in advance and does not require such complex approaches, thus making these mechanisms useless. This means that in the ANN domain the CPU can only be suitable for computing small neural networks, not modern large architectures hundreds of megabytes in extent.

\subsubsection{GPU}

\begin{figure}
  \includegraphics[width=\linewidth]{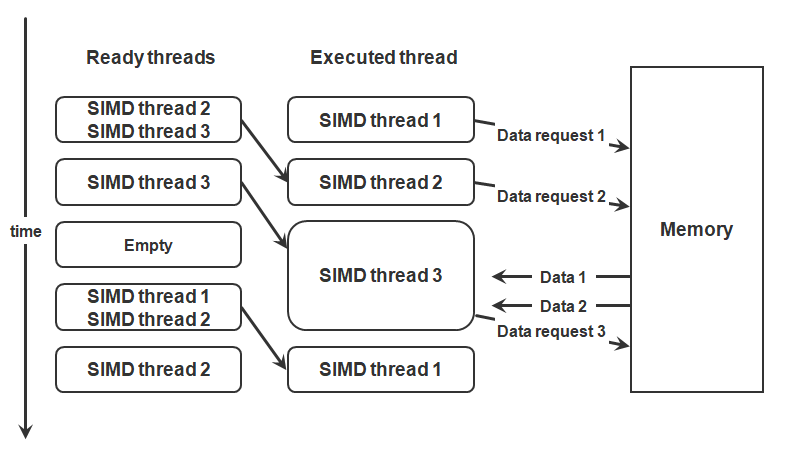}
  \caption{Switching between warps}
  \label{fig:warp_switching}
\end{figure}

The GPU uses several strategies to deal with memory latency. The main one is to give each streaming multiprocessor a large file register that saves the execution context for many threads and provides quick switching between them. The computation scheduler uses this feature and, when an instruction with a high latency is executed in one of the instruction threads (SIMD thread 1), for example, obtaining data from memory, it immediately switches to another instruction thread (SIMD thread 2), and if a latency occurs in there too, the scheduler begins a new ready instruction thread (SIMD thread 3). After some time, data for the first thread arrives and it also becomes ready for execution (see \ref{fig:warp_switching}). This way memory latency can be hidden \cite{hennessy2019computer}. Moreover, in a GPU there can be tens or even more than a hundred such streaming multiprocessors that share the load among themselves.

However, what is crucial aside from latency is the memory throughput, i.e. the maximum amount of data that can be received from the memory per unit time. To solve this problem in GPUs, NVIDIA began adding High Bandwidth Memory (HBM) starting with the P100 (2016), and this dramatically increased their performance compared to previous generations. In the Volta and Turing architectures, Nvidia continued to increase the memory throughput, bringing it up to 1.5 TB/s in the A100 architecture \cite{krashinsky2020nvidia}.

\subsubsection{TPU}

Google announced the first TPU-based processor TPUv1 in 2016 \cite{tpu2018}. It mitigates latency and low memory throughput by using so-called systolic matrices and software-controlled memory instead of caches. The idea of systolic computations is to create a large matrix (256 x 256 for TPUv1) of computing units. Each unit stores a weight and performs two operations. First, it multiplies number x that has come from the unit above by the weight and adds the result to the number that came from the unit to the left. Second, it sends the number x received from above to the unit below, and forwards the received sum to the unit to the right. This is how the TPU performs matrix multiplications in a pipeline. With a sufficiently large batch size, it will not have to constantly access the weights in memory, since the weights are stored in the computing units themselves. At that, having a batch size larger than the width of the systolic array, the TPU will be able to produce one result of multiplying a 256x256 matrix by a 256-long vector each cycle.

Also, unlike the CPU and the GPU that spend energy on constantly accessing registers, transferring values from registers to Arithmetic Logic Unit (ALU) and saving them back to the registers, the TPU can reuse input values and does not have to repeatedly access the registers. This is achieved due to the large number of ALUs in the TPU.

\subsubsection{Neuromorphic Approach}

Despite significant advances and market dominance of the hardware discussed above, the AI systems based on them are still far from their biological counterparts. There is a gap in the level of energy consumption, flexibility (the ability to solve many different tasks), adaptability and scalability. However, such problems are not observed in the mammalian brain. In this connection, it can be assumed that, as it already happened with the principle of massive parallelism  \cite{rumelhart1988parallel}, the implementation of crucial properties and principles of the brain operation could reduce this gap. As a response to this need, a neuromorphic approach to the development of AI systems has appeared in recent years.

The brain is an example of a fundamentally different, non von Neumann, computer. Unlike in classical neural networks executed in modern computing systems, in the brain:

\begin{itemize}

\item Neurons exchange information using discrete impulses - spikes;
\item All events occur asynchronously, there is no single process that explicitly synchronizes the work of all neurons;
\item Learning processes are local and network topologies are non-layered;
\item There is no common memory that universal processors would work with. Instead, a huge number of simple agents function in a self-organizing way.

\end{itemize}

\section{Neuromorphic Approaches in Computing Systems}

Today, many kinds of neural network accelerators are called neuromorphic AI systems to attract attention. To reduce the degree of uncertainty in classifying AI systems as neuromorphic, we propose a list of neuromorphic properties that appear to be useful in creating computing systems and have proven themselves in real-life projects:

\begin{itemize}
\item neural network - based approach: the capability of learning (no need to set the parameters explicitly), the emergence of intellectual properties by linking a large number of relatively simple elements into a network,
\item parallelism: parallel work of neurons, simultaneous execution of different tasks,
\item asynchrony: no single synchronizing process,
\item impulse nature of information transmission: minimum overhead for signal transmission and signal processing at the receiving neuron, resistance to noise,
\item local learning: lower overhead for data transfer operations during learning, the ability to create unlimitedly large systems, continuous and incremental learning,
\item sparsity of data streams: event-driven signal processing, lower overhead for data transfer and data processing,
\item analog computing: efficient hardware implementation.
\item in-memory computing: no overhead for transferring intermediate data, no competitive memory access.

\end{itemize}

Let us consider neuromorphic properties in more detail in the following sections.

\subsection{Parallelism}

Each biological neuron is an independent computer, but much slower than silicon processors. However, the number of biological neurons in the brain that perform coordinated work reaches 87 billion. Back in the late 1980s, researchers \cite{rumelhart1988parallel} came to the conclusion that massively parallel architectures would be required for the efficient operation of neural networks. It was the massive use of the highly parallel architectures (GPU) that had ensured the current success of neural networks.

\subsection{Asynchrony}

However, parallelism by itself does not always give the desired computing effect when synchronization between computing nodes is required. Then, according to Amdahl’s law, the synchronization overheads grow non-linearly as the number of computers increases, thus limiting the gain from parallelism. But the brain seems to have no mechanism that explicitly synchronizes the work of all neurons. Biological neurons work asynchronously, which makes it possible to bypass the limitations of Amdahl’s law.

\subsection{Impulse Nature of Information Transmission}

\begin{figure}
  \includegraphics[width=\linewidth]{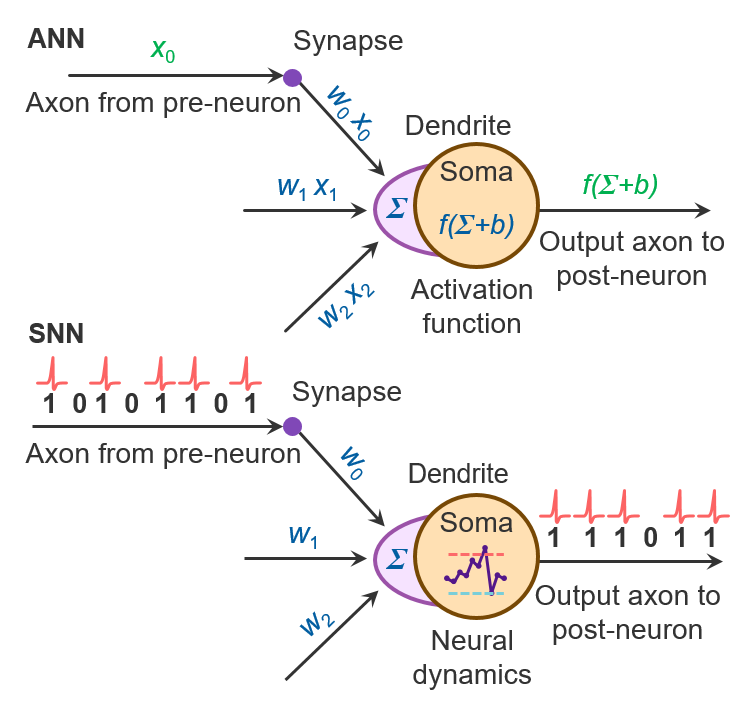}
  \caption{Spiking model vs. classical model \cite{tianjic}.}
  \label{fig:ann_vs_snn}
\end{figure}

In the brain, information is transmitted in the form of nerve impulses, i.e. abrupt, short changes in potential that travel along nerve fibers and always have the same duration and amplitude. Spiking Neural Network (SNN) is a popular mathematical model that describes the impulse nature of information (see \ref{fig:ann_vs_snn}). In SNNs, neurons exchange spikes, i.e. elementary events that have no attributes other than the time of their generation. In that, the transmission of a spike from neuron to neuron does not occur instantly, but requires some time that varies for different pairs of neurons. Thus, each synapse can be characterized not only by the weight w but also by the time delay d. Spike times and delays serve as a mechanism for explicitly introducing time into the computing model.

Information transmission in the form of impulses appears to have key advantages as compared to transmission of real numbers, used in traditional neural networks:

\begin{itemize}
\item data can be transferred between neurons in a simple asynchronous way,
\item SNNs make it possible to work with dynamic data, as it explicitly includes a time component. In SNN, information is encoded based on the time of spike generation and the presence of a delay in spike propagation from neuron to neuron,
\item SNN is a complex non-linear dynamic system,
\item it is energy efficient. The activity of a neuron is reduced to its reaction to the arriving spike, hence after this reaction is complete, the neuron goes into an inactive state that does not require energy. Thus, at each moment of time only a small part of neurons in the network is in the ‘operating mode’ and consumes energy.

\end{itemize}
However, today we see only a few SNN applications in practical tasks. In addition to difficulties with hardware, classical algorithms still outperforms SNN in problem solving quality. Despite a large number of academic papers demonstrating solutions of model problems, SNN training and SNN topologies remain open issues.

\subsection{Local Learning}
The learning of classical neural networks is based on the backpropagation algorithm, which is a special case of the gradient descent method \cite{rumelhart1986learning}. The use of gradient descent methods in the brain is hardly realistic because it would be necessary to bring to each neuron a corrective signal that is computed somewhere based on the results of the network. This means that a feedback system is also required. But even if it were available, it is unclear how two complementary connection systems (direct and reverse) would exchange information about the weight value in them. This problem is called the weight transport problem \cite{grossberg1987weight}.

An alternative to backpropagation is learning methods built upon the principle of locality. The synaptic weight can be modified only on the basis of some activity characteristics of neurons linked by this synapse. In this case, the ideology of Reinforcement Learning (RL) is often used when the correctness of the network decisions is evaluated by delayed signals of reward and punishment.

In the case of SNNs, the laws of synaptic plasticity used for learning are quite diverse, but many of them are modification of the so-called Spike Timing Dependent Plasticity (STDP) \cite{stdp2010}. In the STDP model, the synapses that received spikes shortly before the neuron generated the spike, are strengthened, while the synapses that received spikes after the neuron generated the spike, are weakened. Local learning enables preserving the principle of asynchronous network operation, hence it is applicable to networks of unlimited size. In addition, in many models of synaptic plasticity, it is possible to implement a change in synaptic weights without multiplication operations.

\subsection{Sparsity of Data Streams}
As known \cite{sparsness_brain, brain_silent_segev}, less than 10\% of neurons in the brain are usually active simultaneously. It is very unlike the inference mode of classical neural networks, in which all neurons are involved in computations. This is determined by the following factors.

The first factor is a high similarity of the subsequent frames. The transmission of the changed part of the signal only allows to decrease traffic dramatically, making the data sparse in time (temporal sparsity). For example, in computer vision, instead of transmitting information about each pixel of the image every tick of time, it is possible to transmit only the events of changing the intensity of specific pixels. This approach is used in event-based cameras like Dynamic Vision Sensors (DVS) that can generate an output signal immediately in a spike form.

The second factor is a threshold value of the membrane potential. Below the threshold, the neuron is silent even in the presence of an input signal. The resulting sparsity in data streams is called spatial (spatial sparsity). A similar idea is implemented in the ReLU activation. The significant number of neurons have an output equal to zero but, when computing on the GPU, these zeros will be multiplied anyway just as other numbers.

The third factror is the sparseness of the graph of neural connections. No fully connected layers have been found in the biological brain. Each neuron has a rather limited number of connections (\~5,000). The sparsity of the data flow, conditioned by the network topology, is called structural (structural sparsity). For example, as shown in \cite{frankle2018lottery}, in deep networks it is possible to zero more than 90\% of the weights of connections, while maintaining the network performance.

\subsection{Analog Computing}
The digital representing and processing information uncovers the potential of numerical methods. However, in terms of the number of computational elements, this approach is expensive. An alternative approach is analog circuits. In AI systems, analog circuits can be used for two purposes: modeling the membrane potential dynamics and modeling the synaptic operations. Let us consider them in more detail.

The behavior of biological neurons is usually modeled by a system of differential equations describing the membrane potential dynamics and the operation of ion pumps. In the absence of an analytical solution, the numerical solution of such a system of equations can be very costly. In the brain, neurons do not contain nodes that implement digital computation. They realize their functional with the help of analog computation (membrane potential dynamics). But there are other physical objects that demonstrate similar dynamics (for example, an RC circuit). Thus, a biological neuron can be modeled not only by numerically solved differential equations, but also by using a suitable analog circuit described by such equations. Analog neurons can be 10,000 times faster and more energy efficient \cite{brainscale2017}, and also they naturally support parallelism. The fundamental disadvantage of analog neurons is the impossibility of directly configure and debug them due to the lack of digital memory. Comparing analog implementations with digital ones, we note that analog neurons implement the ‘one neuron – one computer’ principle, while in digital devices one computing unit usually model many neurons by switching the context between them.

Another area where analog circuits are used is the implementation of synaptic operations. For example, the classical model of a neuron requires the computation of Multiply And Accumulate (MAC) operations that take the form: $sum = W_1 * X_1 + ... + W_n * X_n)$. It can be represented as a combination of Ohm’s and Kirchhoff’s laws: $sum = I_1 * R_1 + ... + I_n * R_n$,
where current I plays the role of signal X, and resistance R expresses the value of weight W. In such a scheme, all elements of the multiply–accumulate operation are performed absolutely in parallel in one tick of time.

\subsection{In-Memory Computing}
When performing neural network emulation on the CPU/GPU, one core models a large number of neurons, sequentially switching context between them. This creates significant time and energy overhead for transferring neuron context values to memory and back. Nothing similar is observed in biological neurons.

A biological neuron implements the principle of in-memory computing. A biological neuron is simultaneously a device that stores its state (memory represented by the membrane potential and the strength of synaptic connections), and a device that performs computations. This approach is free of von Neumann’s limitations rooted in physical separation of the shared memory and the processors. The principle of in-memory computing dictates that the memory of a neuron is isolated from the other neurons. This principle implies the ‘one neuron – one computer’ model, which is inherent to analog implementations of a neuron. However, in digital implementations, this approach is too wasteful since there is the possibility of modeling many neurons by one core due to context switching. That is why a hybrid approach is useful in digital implementations. The memory, located physically close to the computing core, is shared by a group of neurons that are modeled by that core (near-memory computing). 

The Static Random Access Memory (SRAM) memory used for such solutions is rather expensive, and this limits the development of SRAM-based chips.

\section{Neuromorphic Projects}
The field of neuromorphic computing is still in its infancy and there is no consensus on what properties should be copied from the brain. Although the development of this field is far from complete, the first steps have already been taken. Next, we will consider existing projects and approaches that can be called neuromorphic based on the classification proposed in the previous section.

\subsection{TrueNorth}

The TrueNorth project \cite{truenorth} (2014, IBM), created under the auspices of the DARPA SyNAPSE program, is the world’s first industrial neuromorphic chip.

The TrueNorth chip is digital, but it does not include general-purpose computational сores. The chip contains 4,096 neural cores, each simulating 256 spiking neurons in real time and contains about 100 Kbits of SRAM memory for storing the state of synapses. A digital data bus is used for communication between neurons and spikes are represented as Address Event Representation (AER) packets containing the identifier of the emitting neuron and the generation time. Multiplication and division are not supported in the digital circuits of TrueNorth neural cores, only addition and subtraction. The functioning of the neural core is not programmed; it is realized in the form of digital operations fixed at the hardware level.

Each neural core has 256 common inputs that can be arbitrarily connected to 256 neurons modeled in one core, i.e. one neuron cannot have more than 256 synapses. Moreover, the weight of each synapse is coded by 2 bits. This means that if neurons have excitatory and inhibitory synapses, the weight of synapses of each kind within one neuron can be only equal to one value. Such primitive coding scheme does not allow any learning algorithm to be realized directly on the chip.

TrueNorth is suitable for the execution of convolutional (CNN) and recurrent neural networks (RNN) \cite{truenorth}, but only in the inference mode. Another hardware platform (most frequently, GPU) should be used for the learning to be followed by the translation of the learned weights into a configuration of TrueNorth neurons.

As an example, the world’s first event-based gesture recognition system was demonstrated at CVPR 2017 \cite{truenorth_gesture}. It consisted of a DVS camera and a TrueNorth chip, capable of recognizing ten gestures with 96.5\% accuracy in 0.1 seconds of gesture demonstration with a consumption of 0.18 W.

A year later, at CVPR 2018 \cite{truenorth_stereo}, the same team presented an event-based stereo vision system, already consisting of two DVS cameras and eight TrueNorth chips, capable of determining the depth of a scene at 2,000 disparity maps per second, while remaining 200 times more energy efficient than other state-of-the-art solutions.

In 2019 \cite{truenorth_10years} they demonstrated a scene-understanding application that detects and classifies multiple objects in high definition aerial video at a throughput exceeding 100 frames per second.

\subsection{Loihi}
The Loihi project (2018, Intel) \cite{loihi} was the first neuromorphic chip with on-chip learning. A Loihi chip includes 128 neural cores, three Pentium processors, four communication modules for AER packet exchange. Each neural core simulates up to 1,024 spiking neurons and contains 128 Kbyte of SRAM to store the state of the synapses. Thus, a chip simulates approximately 128,000 neurons and up to 128,000,000 synapses. The neuron to neuron transmission of all the spikes is guaranteed, and if the flow of spikes becomes too intensive, the system simply slows down.

Synaptic weights can be from 1 to 9 bits and are dynamically modified, making it possible to learn directly on the chip. Besides the weight, the state of each synapse is described by a synaptic delay of up to 6 bits and some variable occupying up to 8 bits, which can be used as an auxiliary variable in the plasticity law. Local learning is realized by the procedure for recalculating the weights using the formula specified when configuring the core. The formula consists only of addition and multiplication operations.

A number of neurocomputers of different capacities have been created on the basis of Loihi. Pohoiki Springs is the most powerful among them. The system includes 768 Loihi chips combined into 24 modules that are positioned on one motherboard, thus simulating 100,000,000 neurons.

More than a hundred scientific groups around the world are using Loihi in research and applied problems \cite{loihi_results_2021}, for example, for recognition and segmentation of images and smells, processing data sequences, realization of a proportional integral differential controller (PID) based on a spiking network, finding the shortest paths in a graph, and others. Some problems are solved by converting the trained classical neural networks into the SNN form. In other projects, SNNs are trained by using surrogate gradient. At last, in several problems, local learning rules are applied. For example, local learning rules are utilized for robotic arm control \cite{loihi_arm} and copter balancing \cite{loihi_pid}.

Intel announced the creation of the second version of the Loihi chip \cite{loihi2} in 2021. One Loihi 2 chip still contains 128 neural cores, simulating 120,000,000 synapses and 1,000,000 programmable (rather than configurable) neurons. The chip is built using Intel4 7nm technology, contains 2.3 billion transistors and has an area of 31 mm2. Another feature of Loihi 2 is 3D multi-chip scaling, i.e. the possibility of combining multiple chips into one system in a 3D (rather than 2D) space, thereby providing lower overheads for communication between the chips.

Loihi 2 realizes a generalized event-based communication model based on local broadcasts and graded spikes (that is, non-binary spikes), in which the spike value is coded by up to 32 bits. In this model, the spikes generated in the system can have amplitude, making it similar to NeuronFlow (considered below).

Alongside Loihi 2, Intel researchers introduced the Lava framework \cite{lava}. It is a cross-platform, open-source framework that offers a new paradigm for describing process-based computing. Lava-has implementations for CPU, GPU and Loihi 2.

\subsection{Tianjic}

The Tianjic project (2019, Tsinghua University) \cite{tianjic} is the first hybrid chip that can work effectively with both ANNs and SNNs. This possibility comes from the reuse of the same parts of circuits for dealing with different types of neural networks. The additional overhead for such versatility is only 3\% of the chip area. Thus, with the Tianjic chip, it is possible to combine architectures of neural networks of different nature (ANN and SNN) within one system.
One Tianjic chip contains 156 neural cores, simulating 40,000 neurons and 10,000,000 synapses. Each core contains 22 Kbyte of SRAM. The digital data bus is used for communication between the cores, and the signals are represented as AER packets.
Scaling is achieved by combining chips into a 2D mesh network. On-chip learning is not supported. The neural network must be pre-trained on another platform (most frequently, GPU) and transferred into the Tianjic configuration to work in the inference mode. Running SNN on Tianjic is 22 times faster and 10,000 times more energy efficient than on GPU. For ANNs, the gains are also significant:

\begin{itemize}
\item LSTM networks are 467 times more energy efficient,
\item MLPs are 723 times more energy efficient and 35 times faster in terms of frame rates,
\item CNNs are 53 times more energy efficient and 101 times faster in terms of frame rates.
\end{itemize}

An example of using just one Tianjic chip to create a bicycle motion control system is presented in \cite{tianjic} , which includes real-time object detection (CNN), object tracking (CANN), voice control (SNN), obstacle avoidance and balance control (MLP). Another SNN, called a Neural State Machine (NSM), was used to integrate neural networks with each other.

\subsection{SpiNNaker}

The SpiNNaker project (2011, The University of Manchester) \cite{spinnaker2014} was the first hardware platform designed exclusively for SNN research. The second generation of the platform SpiNNaker 2 (2018, Dresden University of Technology and The University of Manchester) \cite{spinnaker2021} is being developed as part of the European Human Brain Project.

SpiNNaker is not a chip - it is a massively parallel computer. Its main component is a specially designed microcircuit that has 18 Mbyte of SRAM and 144 ARM M4 microprocessors. These microprocessors have a very limited set of instructions (for example, they do not support division), but they have high performance and low power consumption. The second generation SpiNNaker added support for rate-based DNN, a whole layer of accelerators for numerical operations (exp, log, random, mac, conv2d) and dynamic power management (different voltages and frequencies for different tasks).

Chips are mounted on boards with 56 chips per board. The boards are mounted into racks of 25 in each rack. The racks are combined into cabinets of 10 in each cabinet. All of these make up the SpiNNaker neurocomputer with 106 processors \cite{spinnaker10million} together with the control PC.

The operation of chips within the entire computing system is asynchronous in relation to each other. This gives the entire system more flexibility and scalability but leads to necessity of using AER packets for spike representation. Different communication strategies may be used (multicast, core-to-core, nearest neighbor).

With SpiNNaker, researchers can solve the problem of modeling the biological brain structures. The real-time simulation of a 1 mm2 cortical column (77,000 neurons, 285,000,000 synapses, 0.1 ms time-step) was demonstrated in \cite{spinnaker_cortical}, while the best result of this benchmark on the GPU is two times slower than real time. Thanks to the asynchrony of SpiNNaker, modeling of a 100 mm2 column, instead of a 1 mm2 one, can be achieved by simply increasing the number of computational modules in the system, which is already unattainable for the GPU due to synchronization limitations.

\subsection{BrainScaleS}

The BrainScaleS project (2020, Heidelberg University) \cite{brainscales2021}  is an ASIC device developed as part of the European Human Brain Project. The main idea of BrainScaleS is to emulate the work of spiking neurons applying analog computations. Electronic circuits are used for analog computations. Such electronic circuits are described by the differential equations resembling the equations expressing behavior of membrane potential in biological neurons. One electronic circuit with a resistor and a capacitor corresponds to one biological neuron.

The first version of BrainScaleS was released as early as 2011, but it did not allow on-chip learning. In the second version, several digital processors were added to support local learning (STDP), in addition to the block of analog neurons. The digital data bus is used for communication between neurons using spikes in the form of AER packets. One chip can emulate 512 neurons and 130,000 synapses. The studies \cite{brainscale2017, brainscale_adv_2019} showed that a BrainScaleS neuron could work 10,000 times faster than a biological neuron in the analog implementation. Besides SNN emulation, BrainScaleS can be useful with classical ANNs, performing a matrix-vector multiplication operation in the analog mode.

The main disadvantage of the analog model of a neuron, based on an electrical circuit, is its inflexibility, i.e. the impossibility of changing the neuron model. The relatively large size of the analog neuron is another significant drawback. Works \cite{brainscale_surrogate_grad, brainscale_analog_inf} give examples of applying BrainScaleS to solve the problems of handwritten digit recognition (MNIST), speech recognition by SNN, and also a number of problems in the domain of ANN. For instance, for the spiking MNIST dataset, the classification accuracy was 97.2\% with a latency of 8 µs, a release of 2.4 µJ per image, and a total consumption of 0.2 W for the entire chip connections. The learning was on-chip, but the surrogate gradient methods were used \cite{brainscale_surrogate_grad}. The paper \cite{brainscale_rl} demonstrates the possibility of local learning for BrainScaleS in the Reinforcement Learning tasks using the R-STDP algorithm. The system was trained to control a slider bar in a computer game similar to Atari PingPong.

BrainScaleS is not the only ASIC for simulating analog neurons: the NeuroGrid project \cite{neurogrid} (2009, Stanford University) was based on the same idea. However, it was decided to exclude it from this review because the project seems to have been abandoned (the latest updates were in 2014).

\subsection{NeuronFlow}

The NeuronFlow project \cite{neuronflow_hybrid} (2020, GrAI Matter Labs) presented the GrAIOne chip. The project implements an idea of creating an accelerator to speed up sparse computations and to deal with event-based data. The chip is capable of accelerating both ANN and SNN but it does not support on-chip learning.

GrAIOne contains 196 neural cores simulating 200,704 neurons. Each core contains 1,024 neurons and SRAM for storing the state. The cores communicate via the digital data bus using AER packets.

The term NeuronFlow denotes an architecture with the underlying idea to speed up computations by using a high correlation of frames  in data-flow processing tasks (audio, video). For example, each next frame differs only a little from the previous one in a regular video. Therefore, most neuron activations for the two consequent frames will also be very similar. Then it is possible to avoid sending activation from one neuron to another if it has not changed significantly from what it was at the previous step. This approach gives an opportunity to drastically reduce the number of synaptic operations (multiplications of weights by input values) and memory access operations. Thus, the NeuronFlow architecture is suitable only for processing slowly changing data; otherwise, its advantages are cancelled out.

The paper \cite{sparnet2020} demonstrates the optimization of the PilotNet neural network operation by reducing the number of floating point operations by 16 times. PilotNet is Nvidia’s architecture for controlling the steering wheel of an unmanned vehicle. The network receives an image from the front view camera as an input and calculates the steering wheel angle.

\subsection{DYNAP}

DYNAP (Dynamic Neurormorphic Asynchronous Processors) is a family of solutions from SynSence, a company from the University of Zurich. The company has a patented event-routing technology for communication between the cores.

According to \cite{dynap_routing}, the scalability of neuromorphic systems is mainly limited by the technologies of communication between neurons, and all other limitations are not so important. Researchers at SynSence invented and patented a two-level communication model based on choosing the right balance between point-to-point communication between neuron clusters and broadcast messages within clusters. The company has presented several neuromorphic processors (ASICs): DYNAP-SE2, DYNAP-SEL and DYNAP-CNN.

The Dynap-SE2 and Dynap-SEL chips are not commercial projects and are being developed by neuroscientist as tools for their research. But Dynap-CNN (2021 tinyML) is marketed as a commercial chip for efficient execution of CNNs converted to SNNs. Whereas the Dynap-SE2 and Dynap-SEL research chips implement analog computing and digital communication, Dynap-CNN is fully digital.

Dynap-SE2 is designed for feed-forward, recurrent and reservoir networks. It includes four cores with 1k LIFAT analog spiking neurons and 65k synapses with configurable delay, weight and short term plasticity. There are four types of synapses (NMDA, AMPA, GABAa, GABAb). The chip is used by researches for exploring topologies and communication models of the SNN.

Main distinctive features of Dynap-SEL chip are support for on-chip learning and large fan-in/out network connectivity. It has been created for biologically realistic networks emulation. The Dynap-SEL chip includes five cores, only one of which has plastic synapse. The chip realizes 1,000 analog spiking neurons and up to 80,000 configurable synaptic connections, including 8,000 synapses with integrated spike-based learning rules (STDP). Researchers are using the chip to model cortical networks.

The Dynap-CNN chip has been available with the Development Kit since 2021. Dynap-CNN is a 12 mm2 chip, fabricated in 22nm technology, hosting over one million spiking neurons and four million programmable parameters. Dynap-CNN is completely digital and realizes linear neuron model without leakage. The chip is best combined with event-based sensors (DVS) and is suitable for image classification tasks. In the inference mode the chip can run a SNN converted from a CNN, in which there may be not more than nine convolutional or fully connected layers and not more than 16 output classes. On-chip learning is not supported. The original CNN must be initially created with PyTorch and trained by classical methods (for example, on GPU). Further, using the Sinabs.ai framework (an open source PyTorch based library), the convolutional network can be converted to a spiking form for execution on Dynap-CNN in the inference mode.

Dynap-CNN has demonstrated the following results:

\begin{itemize}
\item CIFAR-10: 1mJ at 90\% accuracy,
\item attention detection: less than 50 ms and 10 mW,
\item gesture recognition: less than 50 ms and 10m W at 89\% accuracy,
\item wake phrase detection: less than 200 ms at 98\% sensitivity and false-alarm rate less than 1 per 100 hours (office background).
\end{itemize}

\subsection{Akida}

Akida \cite{akida} is the first commercial neuromorphic processor, commercially available since August 2021. It has been developed by Australian BrainChip since 2013. Fifteen companies, including NASA, joined the early access program. In addition to Akida System on Chip (SoC), BrainChip also offers licensing of their technologies, providing chip manufacturers a license to build custom solutions.

The chip is marketed as a power efficient event-based processor for Edge computing, not requiring an external CPU. Power consumption for various tasks may range from 100 µW to 300 mW. For example, Akida is capable of processing at 1,000 frames/Watt (compare to TrueNorth with 6,000 frames/Watt). The first generation chip supports operations with convolutional and fully connected networks, with the prospect to add support of LSTM, transformers, capsule networks, recurrent and cortical neural networks. ANN network can be transformed into SNN and executed on the chip.

One Akida chip in a mesh network incorporates 80 Neural Processing Units (NPU), which enables modeling 1,200,000 neurons and 10,000,000,000 synapses. The chip is built at TSMC 28 nm. In 2022, BrainChip announced the second generation chip at 16 nm.

Akida’s ecosystem provides a free chip emulator, TensorFlow compatible framework MetaTF for transformation of convolutional and fully connected neural networks into SNN, аnd a set of pre-trained models. When designing a neural network architecture for execution at Akida, one should take into account a number of additional limitations concerning the layer parameters (e.g. maximum convolution size is 7, while stride 2 is supported for convolution size 3 only) and their sequence.

The major distinctive feature is that incremental, one-shot and continuous learning are supported straight at the chip. At the AI Hardware Summit 2021 BrainChip showed the solution capable of identifying a human in other contexts after having seen him or her only once. Another product by BrainChip is a smart speaker, that on having heard a new voice asks the speaker to identify and after that calls the person by their name. There results are achieved with help of a proprietary local training algorithm on the basis of homeostatic STDP. Only the last fully connected layer supports synaptic plasticity and is involved in learning.

Another instructive case from the AI Hardware Summit 2021 was a classification of fast-moving objects (for example, a race car). Usually, such objects are off the frame center and significantly blurred but they can be detected using an event-based approach.

\section{Memristors}

The neuromorphic hardware systems discussed above are based on the existing complementary metal-oxide-semiconductor (CMOS) technology. There is no direct similarity at the level of physical mechanisms between CMOS devices and elements of biological neural networks, and because of this, CMOS devices can only numerically simulate biological neural networks.

The interest in neuromorphic systems that follow the rules of biological learning, has prompted to explore alternative technologies closer to the biological prototypes. Currently, the most mature technology of this kind is memristors. A memristor is a two-terminal device capable of changing its conductivity depending on the voltage/current applied to the terminals. Such an element was theoretically predicted in 1971, and its practical existence was experimentally demonstrated in 2008. In subsequent years, it was discovered that many materials, mainly at the nanometer scale, can exhibit memristive properties with different physical mechanisms of conductivity switching \cite{camunas2019neuromorphic}.

Now, there are two main directions of memristor usage in neuromorphic applications: vector-matrix multiplication in memory and spiking neural networks.

\subsection{Vector-Matrix Multiplication in Memory}

Main operations of classical neural networks, built upon CMOS technology, are as follows: multiplication, addition and activation function computation. Weights of neural networks are generally stored in SRAM or DRAM cells. CMOS circuits are scalable but the available scalability is still not enough for many neural network applications. Besides, SRAM cell size is too great for high-density integration, while DRAM cells require periodical refreshing to prevent data loss. In neural computation, it is frequently needed to extract data from the memory, transfer data to the computing core, perform computations, and then send the results to the memory through the same data bus. Such an operation sequence being applied to a large amount of data stored in the memory causes a significant computation speed limitation and a large power consumption. This factor substantively limits efficiency of the deep learning technique in the field of big data \cite{xia2019memristive}.

\begin{figure}
  \includegraphics[width=\linewidth]{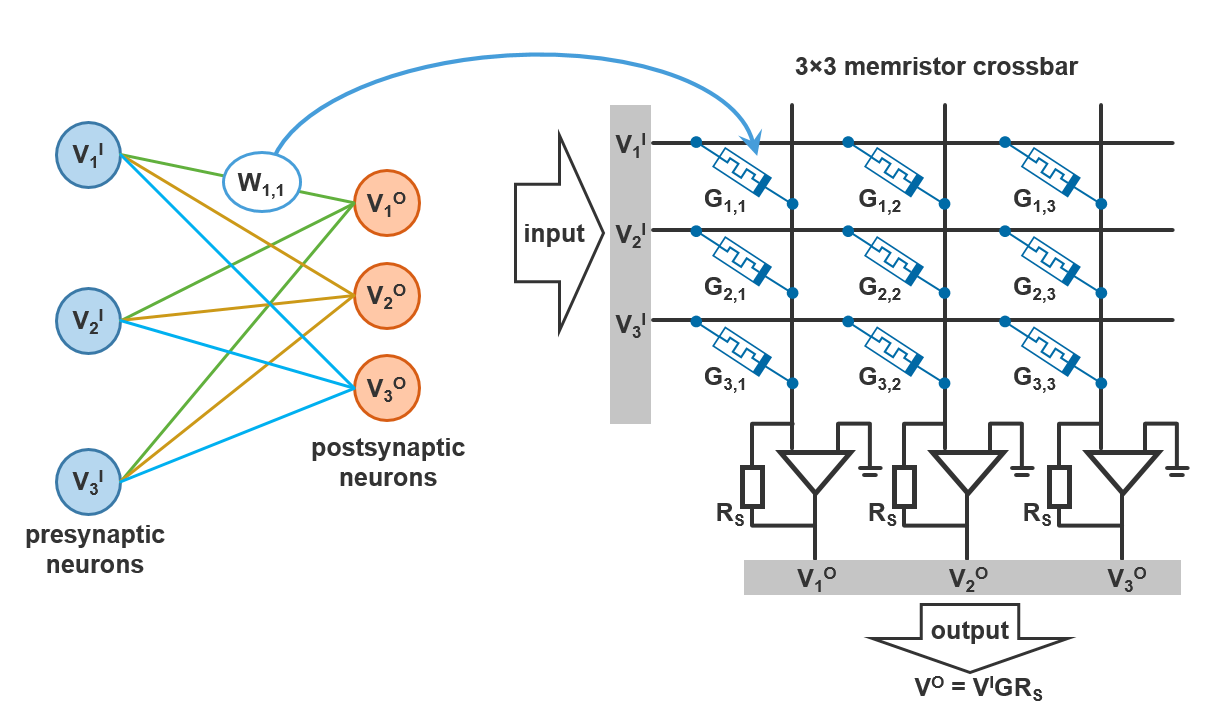}
  \caption{Typical 3×3 memristor crossbar used in neuromorphic applications}
  \label{fig:crossbar}
\end{figure}

Memristor crossbar circuits make it possible to combine addition, multiplication and data storage in a single element. Crossbar is a junction of conducting wires, placed perpendicularly to each other, with memristors positioned at the intersections (see \ref{fig:crossbar}). As it is seen, data are processed and stored in the crossbar. It leads to saving chip space and achieving very low energy consumption.

Researchers have developed various topologies and training algorithms for memristor based neural networks. In the study \cite{hu2014memristor} it was demonstrated a one-layer neural network with a 128×64 memristor massive. During The experiment the 89.9\% accuracy of image recognition for MNIST data set was achieved. In \cite{yao2020fully} researchers demonstrated convolutional neural network (CNN) based on eight 2048 cells memristor arrays The recognition accuracy for MNIST exceeded 96\%. This neuromorphic system outperforms the Tesla V100 110 times in energy efficiency (11 GOPS/W)  and  30 times in performance density(1.2 GOPS/mm2).  The study \cite{li2019long} demonstrated a two-layer recurrent neural network (RNN) based on 14 memristor LSTM blocks. This network achieved 79.1\% accuracy for the task of a human walking classification on the USF-NIST data set

\subsection{Spiking Neural Networks}

Hardware demonstrations of SNNs with the use of memristor devices have mostly focused on the unsupervised learning. Synaptic weights change in accordance with the biologically realistic STDP rule. It was experimentally shown that if appropriate signal forms are used then memristor devices can show the weight adaptation behavior similar to STDP. Apart from demonstrations of singular memristor devices \cite{li2018review}, IBM demonstrated an integrated neuromorphic core with 256 neurons, based on CMOS technology, and 65,536 synapses, based on double-pole memristor. These neurons were capable for on-chip learning with a simplified STDP model. The work of the device was demonstrated on the problem of patterns autoassociation \cite{basu2018low, kim2015nvm}.

Other studies based on selection of switching mechanisms and dynamic parameters of memristors demonstrated different basic neuromorphic principles such as: symmetric and asymmetric plasticity (STDP), spike-rate-dependent plasticity (SRDP), long-term depression (LTD) and long-term potentiation (LTP). Their implementations are described in \cite{basu2018low}. The memristor technologies were also used in hardware implementations for the Hodgkin–Huxley, Morris–Lecar and FitzHugh–Nagumo neuron models. Their implementations are given in \cite{sung2018perspective}.

Speaking of the general properties of memristor materials and structures built on their basis, the following main characteristics valuable for neuromorphic approach should be noted:

\begin{itemize}

\item When a current flows through a memristor, there is a change in its physical structure, which leads to a change in its conductivity. This change of the element itself differs from existing charge-storage-based memory cells (DRAM, SRAM, Flash, etc.) by its significantly longer state retention duration. Based on this property, the development of non-volatile resistive random access memory (ReRAM) is underway, which will have: extended data storage life time (> 10 years), low operating voltage (< 1 V), a large number of rewrite cycles (> $10^{17}$ cycles), low power consumption (10 fJ/bit) \cite{mehonic2020memristors}.
\item Memristors can be used both in fully digital (binary) and analog modes. The manifestation of analog properties is the ability to set a fixed conductivity in a continuous range of values. Using the analog property it is possible to get an element that is able to store information in a multilevel mode of conduction states. The number of states in modern memristive structures reaches 256, which corresponds to 8 bits.
\item The conductivity of the memristor depends on the total value and direction of the current passing through it. This ability allows us to consider the memristor as an element that has a memory of the value of the passed current.
\item Memristor operation timescale may vary from second to nanosecond.
\item Memristors can be scaled down to less than 10 nm and made compatible with existing CMOS technology to achieve high computational density \cite{zahoor2020resistive}.

\end{itemize}

Despite the fact that memristor technology demonstrates neuromorphic properties at the elemental level and offers great prospects for a wide range of applications, at the moment the main researchers of memristors remain academic laboratory centers. In order to make memristor technologies commercially viable several serious problems must be solved. These problems include: scatter of parameters manufactured by memristors; non-linearity of current-voltage characteristics; limited conductivity range \cite{im2020memristive}.

\section{Conclusion}

To consolidate all the approaches above, we have prepared the summary tables. Tables \ref{table:final-table-synapses} and \ref{table:final-table-somas} show different approaches to synapses and neuron modeling. Table \ref{table:final-table} provides comparison of all the chips reviewed in this article.

\begin{table}
\begin{tabular}{ |p{45mm}|p{15mm}|p{25mm}|p{25mm}|p{45mm}| } 
\hline
Approach & Network & In-memory computation &  Signal & On-device training  \\
\hline
\multirow{2}{45mm}{Computational modeling on digital logic} & ANN & \multirow{2}{25mm}{no / near-memory} & \multirow{2}{25mm}{digital} & Backprop \\
& SNN & & & Surrogate Gradient / STDP \\
\hline
\multirow{2}{45mm}{Memristors} & ANN & \multirow{2}{25mm}{yes} & \multirow{2}{25mm}{analog} & - \\
& SNN & & & STDP \\
\hline

\end{tabular}
\caption{A comparison of neuromorphic approaches to synapses modeling}
\label{table:final-table-synapses}
\end{table}

\begin{table}
\begin{tabular}{ |p{45mm}|p{15mm}|p{25mm}|p{25mm}| } 
\hline
Approach & Network & In-memory computation &  Signal  \\
\hline
\multirow{2}{45mm}{Computational modeling on digital logic} & ANN & \multirow{2}{25mm}{no / near-memory} & \multirow{2}{25mm}{digital} \\
& SNN & &  \\
\hline
RC circuit & SNN & yes & analog \\
\hline

\end{tabular}
\caption{A comparison of neuromorphic approaches to neuron soma modeling}
\label{table:final-table-somas}
\end{table}

\newpage
\begin{sidewaystable}
\begin{tabular}{ |p{20mm}|p{20mm}|p{20mm}|p{20mm}|p{15mm}|p{10mm}|p{10mm}|p{30mm}|p{50mm}| } 
\hline
Chip / Neural computer  & In-memory computation & Signal & On-device training & Analog & Event-based & nm & Power consumption & Features  \\
\hline
 CPU / GPU / TPU & no & real numbers, spikes & Backprop / STDP & no & no & 5 & Google Edge TPU: 2 Tops/watt & high popularity, rich ecosystem, advanced engineering technologies \\ 
\hline
TrueNorth & near-memory & spikes & no & no & yes & 28 & 400 GOPS/watt 25 pj/operation & first industrial neuromorphic chip without training (IBM) ??? \\
\hline
Loihi & near-memory & spikes & STDP & no & yes & 14 & 80 pj/operation & first neuromorphic chip with training (Intel) \\
\hline
Loihi2 & near-memory & real numbers, spikes & STDP, surrogate backprop & no & yes & Intel4 (7 nm) & - & development of Loihi ideas, addition of non-binary spike support, neurons can be programmed \\
\hline
Tianjic & near-memory & real numbers, spikes & no & no & yes & 28 & 1278 MACGOPS / watt 649 GSOPS/watt & hybrid chip with effective support of both SNN and ANN, energy efficiency \\
\hline
SpiNNaker & near-memory & real numbers, spikes & STDP & no & no & 22 & 20 nj/operation & scalable computer for SNN simulation \\
\hline
BrainScaleS & yes & real numbers, spikes & STDP, Surrogate gradient & yes, мембрана & yes & 65 & 10 pJ/synaptic event, 200 mW - total & analog neurons at RC circuits, large size \\
\hline
GrAIOne (NeuronFlow) & near-memory & real numbers, spikes & no & no & yes & 28 & 20 pj/operation 10 pj/operation & NeuronFlow architecture, effective support of sparse computations, support of ANN and SNN \\
\hline
DYNAP SE2, SEL, CNN & near-memory & spikes & STDP (SEL) & SE2, SEL & yes & 22 & CIFAR-10: 1 mlJ/ Image at 90 \% acc. & proprietary communication protocol \\
\hline
Akida & near-memory & spikes & STDP (last layer) & no & yes & 28 & 100 microwatts – 300 milliwatts 2000 frames/watt & first commercial neuromorphic processor with incremental, one-shot and continuous learning for CNN \\
\hline
\end{tabular}
\caption{Comparison of neuromorphic chips}
\label{table:final-table}
\end{sidewaystable}

\clearpage
\newpage
\printbibliography

\end{document}